# An improvement direction for filter selection techniques using information theory measures and quadratic optimization

Waad Bouaguel
LARODEC, ISG, University of Tunis
41, rue de la Liberté, 2000 Le Bardo, Tunisie.

Ghazi Bel Mufti
ESSEC, University of Tunis
4, rue Abou Zakaria El Hafsi, 1089 Montfleury, Tunisie.

*Abstract*— Filter selection techniques are known for their simplicity and efficiency. However this kind of methods doesn't take into consideration the features inter-redundancy. Consequently the un-removed redundant features remain in the final classification model, giving lower generalization performance. In this paper we propose to use a mathematical optimization method that reduces inter-features redundancy and maximize relevance between each feature and the target variable.

## I. INTRODUCTION

In many classification problems we deal with huge datasets, which likely contain not only many observations, but also a large number of variables. Some variables may be redundant or irrelevant to the classification task. As far as the number of variables increase, the dimensions of data amplify, yielding worse classification performance. In fact with so many irrelevant and redundant features, most classification algorithms suffer from extensive computation time, possible decrease in model accuracy and increase of overfitting risks [17, 12]. As a result, it is necessary to perform dimensionality reduction on the original data by removing those irrelevant features.

Two famous special forms of dimensionality reduction exist. The first one is feature extraction, in this category the input data is transformed into a reduced representation set of features, so new attributes are generated from the initial ones. The Second category is feature selection. In this category a subset of the existing features without a transformation is selected for classification task. Generally feature selection is chosen over feature extraction because it conserves all information about the importance of each single feature while in feature extraction the obtained variables are, usually, not interpretable. In this case it is obvious that we will study the feature selection but choosing the most effective feature selection method is not an easy task.

Many empirical studies show that manipulating few variables leads certainly to have reliable and better understandable models without irrelevant, redundant and noisy data [21, 20]. Feature selection algorithms can be roughly categorized into the following three types, each with different evaluation criteria [7]: filter model, wrapper model and embedded. According to [18, 3, 9] a filter method is a pre-selection process in which a subset of features is firstly selected independently of the later applied classifier. Wrapper method on the other hand, uses search techniques to rank the discriminative power of all of the possible feature subsets and evaluate each subsets based on classification accuracy [16], using the classifier that was incorporated in the feature selection process [15, 14]. The wrapper model generally performs well, but has high computational cost. Embedded method [20] incorporates the feature selection process in the classifier objective function or algorithm. As result the embedded approach is considered as the natural ability of a classification algorithm; which means that the feature selection take place naturally as a part of classification algorithm. Since the embedded approach is algorithm-specific, it is not an adequate one for our requirement.

Wrappers on other hand have many merits that lie in the interaction between the feature selection and the classifier. Furthermore, in this method, the bias of both feature selection algorithm and learning algorithm are equal as the later is used to assess the goodness of the subset considered. However, the main drawback of these methods is the computational weight. In fact, as the number of features grows, the number of subsets to be evaluated grows exponentially. So, the learning algorithm needs to be called too many times. Therefore, performing a wrapper method becomes very expensive computationally.

According to [2, 17] filter methods are often preferable to other selection methods because of their usability with alternative classifiers and their simplicity. Although filter algorithms often score variables separately from each other without considering the inter-feature redundancy, as result they do not always achieve the goal of finding combinations of variables that give the best classification performance [13].Therefore, one common step up for filter methods is to consider dependencies and relevance among variables. mRMR [8] (Minimal-Redundancy-Maximum-Relevance) is an effective approach based on studying the mutual information among features and the target variable; and taking into account the inter-features dependency [19]. This approach selects those features that have highest relevance to the target class with the minimum inter-features redundancy. The mRMR algorithm, selects features greedily.

The new approach proposed in this paper aims to show how using mathematical methods improves current results. We use quadratic programming [1] in this paper, the studied objective function represents inter-features redundancy through quadratic term and the relationship between each feature and the class label is represented through linear term. This work has the following sections: in section 2 we review studies related to filter methods; and we study the mRMR feature selection approach. In section 3 we propose an advanced approach using mathematical programming and mRMR algorithm background. In section 4 we introduce the used similarity measure. Section 5 is dedicated to empirical results.

## II. FILTER METHODS

The processing, of filter methods at most cases can be described as it follows: At first, we must evaluate the features relevance by looking at the intrinsic properties of the data. Then, we compute relevance score for each attribute and we remove ones which have low scores. Finally, the set of kept features forms the input of the classification algorithm. In spite of the numerous advantages of filters, scoring variables separately from each other is a serious limit for this kind of techniques. In fact when variables are scored individually they do not always achieve the object of finding the perfect features combination that lead to the optimal classification performance [13].

Filter methods fail in considering the inter-feature redundancy. In general, filter methods select the top-ranked features. So far, the number of retained features is set by users using experiments. The limit of this ranking approach is that the features could be correlated among themselves. Many studies showed that combining a highly ranked feature for the classification task with another highly ranked feature for the same task often does not give a great feature set for classification. The raison behind this limit is redundancy in the selected feature set; redundancy is caused by the high correlation between features.

The main issue with redundancy is that with many redundant features the final result will not be easy to interpret by business managers because of the complex representation of the target variable characteristics. With numerous mutually highly correlated features the true representative features will be consequently much fewer. According to [8], because features are selected according to their discriminative powers, they do not fully represent the original space covered by the entire dataset. The feature set may correspond to several dominant characteristics of the target variable, but these could still be fine regions of the relevant space which may cause a lack in generalization ability of the feature set.

### A. mRMR Algorithm

A step up for filter methods is to consider dissimilarity among features in order to minimize feature redundancy. The set of selected features should be maximally different from each other. Let S denote the subset of features that we are looking for. The minimum redundancy condition is

$$MinP_1, P_1 = \frac{1}{|S|^2} \sum_{x_i, x_j \in S} M(x_i, x_j), \quad (1)$$

where we use $M(i, j)$ to represent similarity between features, and $/S/$ is the number of features in S. In general, minimizing only redundancy is not enough sufficient to have a great performance, so the minimum redundancy criteria should be supplemented by maximizing relevance between the target variable and others explicative variables. To measure the level of discriminant powers of features when they are differentially expressed for different target classes, again a similarity measure $M(y, x_i)$ is used, between targeted classes y={0,1} and the feature expression $x_i$. This mesure quantifies the relevance of $x_i$ for the classification task. Thus the maximum relevance condition is to maximize the total relevance of all features in S:

$$MaxP_2, P_2 = \frac{1}{|S|^2} \sum_{x_i \in S} M(y, x_i). \quad (2)$$

Combining criteria such as: maximal relevance with the target variable and minimum redundancy between features is called the minimum redundancy-maximum relevance (mRMR) approach. The mRMR feature set is obtained by optimizing the problems $P_1$ and $P_2$ receptively in Eq. (1) and Eq. (2) simultaneously. Optimization of both conditions requires combining them into a single criterion function

$$Min\{P_1 - P_2\}. \quad (3)$$

mRMR approach is advantageous of other filter techniques. In fact with this approach we can get a more representative feature set of the target variable which increases the generalization capacity of the chosen feature set. Consistently, mRMR approach gives a smaller feature set which effectively cover the same space as a larger conventional feature set does. mRMR criterion is also another version of MaxDep [19] that chooses a subset of features with both minimum redundancy and maximum relevance.

In spite of the numerous advantages of mRMR approach; given the prohibitive cost of considering all possible subsets of features, the mRMR algorithm selects features greedily, minimizing their redundancy with features chosen in previous steps and maximizing their relevance to the class. A greedy algorithm is an algorithm that follows the problem solving heuristic of making the locally optimal choice at each stage with the hope of finding a global optimum; the problem with this kind of algorithms is that in some cases, a greedy strategy do not always produce an optimal solution, but nonetheless a greedy heuristic may yield locally optimal solutions that approximate a global optimal solution.. On the other hand, this approach treats the two conditions equally important. Although, depending on the learning problem, the two conditions can have different relative purposes in the objective function, so a coefficient balancing the MaxDep and the MinRev criteria should be added to mRMR objective function. To improve the theory of mRMR approach we use in the next section; mathematical knowledge to modify and balance the mRMR objective function and solve it with quadratic programming.

III. QUADRATIC PROGRAMMING FOR FEATURE SELECTION

*A. Problem Statement*

The problem of feature selection was addressed by statistics machine learning as well as by other mathematical formulation. Mathematical programming based approaches have been proven to be excellent in terms of classification accuracy for a wide range of applications [5, 6]. The used mathematical method is a new quadratic programming formulation. Quadratic optimization process, use an objective function with quadratic and linear terms. Here, the quadratic term presents the similarity among each pair of variables, whereas the linear term captures the correlation of each feature and the target variable.

Assume the classifier learning problem involves $N$ training samples and $m$ variables [20]. A quadratic programming problem aims to minimize a multivariate quadratic function subject to linear constraints as follows:

$$Minf(\mathbf{x}) = \frac{1}{2}\mathbf{x}^T Q\mathbf{x} - F^T\mathbf{x}.$$

*Subject to*

$$\begin{cases} x_i \geq 0 \forall i = 1, \text{K}, m \\ \sum_{i=1}^{m} x_i = 1. \end{cases} \quad (4)$$

where $F$ is an m-dimensional row vector with non-negative entries, describing the coefficients of the linear terms in the objective function. $F$ measures how correlated each feature is with the target class (relevance). $Q$ is an $(m \times m)$ symmetric positive semi-definite matrix describing the coefficients of the quadratic terms, and represents the similarity among variables (redundancy). The weight of each feature decision variables are denoted by the m-dimensional column vector $\mathbf{x}$.

We assume that a feasible solution exists and that the constraint region is bounded. When the objective function $f(\mathbf{x})$ is strictly convex for all feasible points the problem has a unique local minimum which is also the global minimum. The conditions for solving quadratic programming, including the Lagrangian function and the Karush-Kuhn-Tucker conditions are explained in details in [1]. After the quadratic programming optimization problem has been solved, the features with higher weights are better variables to use for subsequent classifier training.

*B. Conditions balancing*

Depending on the learning problem, the two conditions can have different relative purposes in the objective function. Therefore, we introduce a scalar parameter $\alpha$ as follows:

$$Minf(\mathbf{x}) = \frac{1}{2}(1-\alpha)\mathbf{x}^T Q\mathbf{x} - \alpha F^T\mathbf{x}, \quad (5)$$

above $\mathbf{x}$, $Q$ and $F$ are defined as before and $\alpha \in [0,1]$, if $\alpha = 1$, only relevance is considered. On the opposing, if $\alpha = 0$, then only independence between features is considered that is, features with higher weights are those which have lower similarity coefficients with the rest of features. Every data set has its best choice of the scalar $\alpha$. However, a reasonable choice of $\alpha$ must balances the relation between relevance and redundancy. Thus, a good estimation of $\alpha$ must be calculated. We know that the relevance and redundancy terms in Equation 6 are balanced when $(1-\alpha)\overline{Q} = \alpha\overline{F}$, where $\overline{Q}$ is the estimate of the mean value of the matrix $Q$; and $\overline{F}$ is the estimate of the mean value of vector $F$ elements. A practical estimate of is defined as

$$\hat{\alpha} = \frac{\overline{Q}}{\overline{Q} + \overline{F}}. \quad (6)$$

IV. BASED INFORMATION THEORY SIMILARITY MEASURE

The information theory approach has proved to be effective in solving many problems. One of these problems is feature selection where information theory basics can be exploited as metrics or as optimization criteria. Such is the case of this paper, where we exploit the mean value of the mutual information between each pair of variables in the subset as metric in order to approximate the similarity among features. Formally, the mutual information of two discrete random variables $x_i$ and $x_j$ can be defined as:

$$I(x_i, x_j) = \sum_{x_i \in S} \sum_{x_j \in S} p(x_i, x_j) log \frac{p(x_i, x_j)}{p(x_i)p(x_j)}, \quad (7)$$

and of two continuous random variables is denoted as follows:

$$I(x_i, x_j) = \iint p(x_i, x_j) log \frac{p(x_i, x_j)}{p(x_i)p(x_j)} dx_i dx_j. \quad (8)$$

V. 5. EMPIRICAL STUDY

In general mutual information computation requires estimating density functions for continuous variables. For simplicity, each variable is discretized using Weka 3.7 software [4]. We implemented our approach in R using the quadprog package [10, 11]. The studied approach should be able to give good results with any classifier learning algorithms, for simplicity the logistic regression provided by R will be the underlying classifier in all experiments.

The generality of the feature selection problem makes it applicable to a very wide range of domains. We chose in this paper to test the new approach on two real word credit scoring datasets from the UCI Machine Learning Repository.

The first dataset is the German credit data set consists of a set of loans given to a total of 1000 applicants, consisting of 700 examples of creditworthy applicants and 300 examples where credit should not be extended. For each applicant, 20 variables describe credit history, account balances, loan purpose, loan amount, employment status, and personal information. Each sample contains 13 categorical, 3 continuous, 4 binary features, and 1 binary class feature. The second data set is the Australian credit dataset which is composed by 690 instances where 306 instances are creditworthy and 383 are not. All attributes names and values have been changed to meaningless symbols for confidential reason. Australian dataset present an interesting mixture of continuous features with small numbers of values, and nominal with larger numbers of values. There are also a few missing values.

The aim of this section is to compare classification accuracy achieved with the quadratic approach and others filter techniques. Table I and Table ii show the average classification error rates for the two data sets as a function of the number of features. Accuracy results are obtained with α= 0.511 for German data set and α= 0.489 for Australian data set, which means that an equal tradeoff between relevance and redundancy is best for the two data sets. From Table 1 and Table II it's obvious that using the quadratic approach for variable selection lead to the lowest error rate.

TABLE I. RESULTS SUMMARY FOR GERMAN DATASET, WITH 7 SELECTED FEATURES

|  | Test error | Type I error | Type II error |
|---|---|---|---|
| Quadratic | 0.231 | 0.212 | 0.222 |
| Relief | 0.242 | 0.233 | 0.287 |
| Information Gain | 0.25 | 0.238 | 0.312 |
| CFS Feature Set Evaluation | 0.254 | 0.234 | 0.344 |
| mRMR | 0.266 | 0.25 | 0.355 |
| MaxRel | 0.25 | 0.238 | 0.312 |

TABLE II. RESULTS SUMMARY FOR AUSTRALIAN DATASET, WITH 6 SELECTED FEATURES

|  | Test error | Type I error | Type II error |
|---|---|---|---|
| Quadratic | 0.126 | 0.155 | 0.092 |
| Relief | 0.130 | 0.164 | 0.099 |
| Information Gain | 0.127 | 0.163 | 0.094 |
| CFS Feature Set Evaluation | 0.126 | 0.165 | 0.098 |
| mRMR | 0.130 | 0.164 | 0.099 |
| MaxRel | 0.139 | 0.179 | 0.101 |

VI. CONCLUSION

This paper has studied a new feature selection method based on mathematical programming; this method is based on the optimization of a quadratic function using the mutual information measure in order to capture the similarity and nonlinear dependencies among data.

ACKNOWLEDGMENT

The authors would like to thank Prof. Mohamed Limam who provides valuable advices, support and guidance. The product of this research paper would not be possible without his help.

REFERENCES

[1] M. Bazaraa, H. Sherali, C. Shetty, Nonlinear Programming Theory and Algorithms, JohnWiley, New York, 1993.
[2] R. Bekkerman, E. Yaniv r, N. Tishby, Y.Winter, "Distributional word clusters vs. words for text categorization", J. Mach. Learn. Res., vol. 3, 2003, pp. 1183–1208.
[3] A. L. Blum, P. Langley, "Selection of relevant features and examples in machine learning", Artificial intelligence, vol. 97, 1997, pp. 245–271.
[4] R. R. Bouckaert, E. Frank, M.Hall, R. Kirkby, P. Reutemann, A. Seewald, D. Scuse, "Weka manual (3.7.1) ", 2009.
[5] P. S. Bradley, O. L. Mangasarian, W. N. Street, "Feature Selection Via Mathematical Programming", INFORMS J. on Computing, vol. 10, 1998, pp. 209–217.
[6] P. S. Bradley, U. M. Fayyad, Mangasarian, "Mathematical Programming for Data Mining : Formulations and Challenges", INFORMS J. on Computing, vol. 11, num. 3, 1999, p. 217–238, INFORMS.
[7] Y.S. Chen, "Classifying credit ratings for Asian banks using integrating feature selection and the CPDA-based rough sets approach", Knowledge-Based Systems, 2011.
[8] C. Ding, H. Peng, "Minimum Redundancy Feature Selection from Microarray Gene Expression Data", J. Bioinformatics and Computational Biology, vol. 3, num. 2, 2005, pp. 185-206.
[9] G. Forman, "BNS feature scaling : an improved representation over tf-idf for svm text classification", CIKM'08: Proceeding of the 17th ACM conference on Information and knowledge mining, New York, NY, USA, 2008, ACM, pp. 263–270.
[10] D. Goldfarb, A. Idnani, "Dual and Primal-Dual Methods for Solving Strictly Convex Quadratic Programs", In J. P. Hennart (ed.), Numerical Analysis, 1982, pp. 226–239, Springer Verlag.
[11] D. Goldfarb, A. Idnani, "A numerically stable dual method for solving strictly convex quadratic programs. ", Mathematical Programming, , 1983, pp. 1–33.
[12] T. Howley, M.G. Madden, M. L. O'connell, A.G. Ryder, "The effect of principal component analysis on machine learning accuracy with high-dimensional spectral data. ", Knowl.-Based Syst., vol. 19, num. 5, 2006, pp. 363-370.
[13] A. K. Jain, R. P. W. Duin, J. Mao, "Statistical pattern recognition : a review", Pattern Analysis and Machine Intelligence, IEEE Transactions on, vol. 22, num. 1, 2000, pp. 4–37, IEEE.
[14] R. Kohavi, G. H. John, "Wrappers for Feature Subset Selection", Artificial Intelligence, vol. 97, num. 1, 1997, pp. 273–324.
[15] D. Koller, M. Sahami, "Toward Optimal Feature Selection", International Conference on Machine Learning, 1996, pp. 284–292.
[16] S. Yuan kung, "Feature selection for pairwise scoring kernels with applications to protein subcellular localization", in IEEE Int. Conf. on Acoustic, Speech and Signal Processing (ICASSP), 2007, pp. 569–572.
[17] Y. liu, M. Schumann, "Data mining feature selection for credit scoring models", Journal of the Operational Research Society, vol. 56, num. 9, 2005, pp. 1099–1108.
[18] L.C. Molina, L. Belanche, A. Nebot, "Feature Selection Algorithms : A Survey and Experimental Evaluation", Data Mining, IEEE International Conference on, vol. 0, 2002, pp 306, IEEE Computer Society.
[19] H. Peng, F. Long, C. Ding, "Feature selection based on mutual information: criteria of maxdependency, maxrelevance, and min-redundancy", IEEE Transactions on Pattern Analysis and Machine Intelligence, vol. 27, 2005, pp. 1226–1238.
[20] I. Rodriguez-lujan , R. Huerta, C. Elkan, C. S. Cruz, "Quadratic Programming Feature Selection", Journal of Machine Learning Research, vol. 11, 2010, pp. 1491–1516.
[21] C. M.Wang, W. F. Huang, "Evolutionary-based feature selection approaches with new criteria for data mining: A case study of credit approval data", Expert Syst. Appl., vol. 36, num. 3, 2009, pp. 5900-5908.